# Features Based Text Similarity Detection

*Chow Kok Kent, Naomie Salim*

*Faculty of Computer Science and Information Systems, University Teknologi Malaysia, 81310 Skudai, Johor, Malaysia*

**Abstract**— As the Internet help us cross cultural border by providing different information, plagiarism issue is bound to arise. As a result, plagiarism detection becomes more demanding in overcoming this issue. Different plagiarism detection tools have been developed based on various detection techniques. Nowadays, fingerprint matching technique plays an important role in those detection tools. However, in handling some large content articles, there are some weaknesses in fingerprint matching technique especially in space and time consumption issue. In this paper, we propose a new approach to detect plagiarism which integrates the use of fingerprint matching technique with four key features to assist in the detection process. These proposed features are capable to choose the main point or key sentence in the articles to be compared. Those selected sentence will be undergo the fingerprint matching process in order to detect the similarity between the sentences. Hence, time and space usage for the comparison process is reduced without affecting the effectiveness of the plagiarism detection.

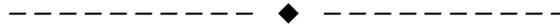

## 1 INTRODUCTION

In plagiarism detection, the content of a suspected document may be represented as a collection of terms, words, stems, phrases, or other units derived or inferred from the text of the document. Different techniques will lead to vary efficiency and effectiveness in plagiarism detection based on different document descriptors. Before a document is taken to be compared, it is necessary for us to choose the most appropriate representation techniques to retrieve the main points of the document. When the documents contain primarily unrestricted text such as newspaper articles, legal documents and so on, the relevance of a document is established through 'full-text' retrieval. This has been usually accomplished by identifying key terms in the documents. There are a few techniques that have been developed or adapted for plagiarism detection in natural language documents. The most common technique used nowadays is the Fingerprint Matching technique [1][2]that consists of the process of scanning and examining the fingerprint of two documents in order to detect plagiarism.

## 2 FINGERPRINT MATCHING TECHNIQUE

Fingerprinting techniques mostly rely on the use of K-grams (Manuel et al. 2006) because the process of fingerprinting divides the document into grams of certain length k. Then, the fingerprints of two documents can be compared in order to detect plagiarism. It has been observed through the literature that fingerprints matching approach differs based on what representation or comparison unit (i.e.grams) is used.

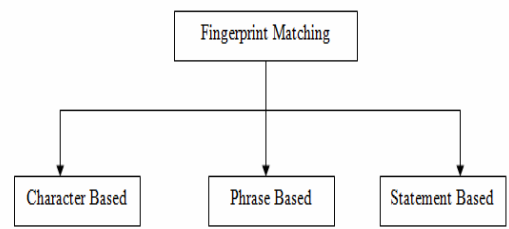

Fig.1 Fingerprint Matching Technique

### 2.1 Character-based Fingerprint Matching

The conventional fingerprinting technique uses sequence of characters to form the fingerprint for the whole document. During 1996, Heintze divides fingerprinting techniques into two types which are full and selective. In full fingerprinting, document fingerprint consists of the set of all possible substrings of length K. For example, if we have a document of length $|D| = 5$ consisting only one statement that has only one word "touch", then we can see that "touc" and "ouch" are the all possible substrings of length K = 4. In general, there are $|D| – k + 1$ substrings or k-grams, where $|D|$ is the length of the document. Basically, comparing two documents under this technique is counting the number of substrings that are common in both fingerprints [1].

Hence, if we compare a document A of size |A| against a document B, and if N is the number of substrings common in both, then the resemblance measure R of how much of A is contained in B can be computed as follows:

$$R = \frac{N}{|A|} \quad \text{where} \quad 0 <= R <= 1$$

It is important to choose the right value of k to provide good discrimination among documents. If the value of k is chosen appropriately, then full fingerprinting gives reliable exact match results. The value given by Heintze (1996) was effectively 30-45 characters. Although full fingerprinting can be considered as a time and space consuming burden, it is a very useful measure for document copy detection.

## 2.2 Phrase-based Fingerprint Matching

In 2001, Lyon et al. generates fingerprint using phrase-mechanism to measure the resemblance between two documents. During the early stage, we have to convert each document to a set of trigrams (three words). Hence, a sentence such as *"Web Based Cross Language Plagiarism Detection"* will be converted to the set trigrams {*"Web Based Cross"*, *"Based Cross Language"*, *"Cross Language Plagiarism"*, *"Language Plagiarism Detection"*}. Then, the set of trigrams for each document is compared with all other using the matching algorithm. Finally, the measure of the resemblance for each pair of documents is calculated as follows:

$$R = \frac{S(A) \cap S(B)}{S(A) \cup S(B)} \quad \text{where} \quad 0 <= R <= 1$$

where S(A) and S(B) are the set of all trigrams in documents A and B, respectively. When the value of R is approaching 1, it means that the document is identical to the respective pair of document. Phrase-based fingerprint matching will works better and faster than character-based fingerprinting since it deals with words rather than letters.

## 2.3 Statement-based Fingerprint Matching

The pros and cons of character-based and phrase-based fingerprinting have led Yerra and Ng (2005) to represent the fingerprints of each statement (and thence the whole document) by three least-frequent 4-grams. Although any value of K can be considered, yet K = 4 was stated as an ideal choice by Yerra and Ng (2005). This is because smaller values of K (i.e., K = 1, 2, or 3), do not provide good discrimination between sentences. On the other hand, the larger the values of K (i.e., K = 5, 6, 7...etc), the better discrimination of words in one sentence from words in another. However each K-gram requires K bytes of storage and hence space-consuming becomes too large for larger values of K. Therefore, we can conclude that K = 4 is an optimal or near optimal choice. Here is an explanation of how this 3-least frequent 4- grams works. A 4-gram of a string is a set of all possible 4-character substrings. For example, let take a string S = "English Word", then the possible set of 4-grams include *"engl, ngli, glis, lish, ishw, shwo, hwor, word"* with ignoring spaces.

Secondly, three least-frequent 4-grams are the best option to represent thesentence uniquely. To illustrate the three least-frequent 4-gram construction process, consider the following sentence S *"soccer game is fantastic"*. The 4-grams are *socc, occe, ccer, cerg,* etc. In this method, instead of comparing all possible 4-grams, only three 4-grams which have the least frequency over all 4-grams will be chosen. Gauge the frequency or the weight of each n-gram was stated by as follows [3].

Let the document contain J distinct n-grams, with mi occurrences of n-gram number
    i. Then the weight assigned to the i[th] n-gram will

$$xi = \frac{mi}{\sum_{j=1}^{J} mj}$$

Where

$$\sum_{i=1}^{J} xi = 1$$

Thirdly, the three least-frequent 4-grams are concatenated to represent the fingerprint of a sentence from the query document to be compared with the three least-frequent 4-gram representations of sentences in the corpora documents. Thus, in our example, if the three least-frequent 4-grams from Snew are *occe, ccer, cerg* will be concatenated to form the fingerprint F of Snew *"occeccercerg"*. Finally, two sentences are treated the same if their corresponding three least frequent 4-gram representations are the same. A measure of resemblance for each pair of documents is computed as follows:

$$R = \frac{F(A) \cap F(B)}{F(A) \cup F(B)} \quad 0 <= R <=1$$

where F(A) and F(B) are the common fingerprints in documents A and B, respectively.

The main advantages of the statement-based fingerprint method are the time and space consumption issue. The processing time and space usage for this method are much better than the character-based and phrase-based techniques. However, in the case of rewording and restructuring of statements, all fingerprinting techniques will fail in detecting that kind of plagiarism.

## 3 FEATURES BASED TEXT SIMILARITY DETECTION

In previous section, we discuss the implementation of the fingerprint matching technique in the plagiarism detection tools. It cannot be denied that those matching techniques play an essential role in detecting the plagiarism documents. However, there are still some weaknesses in these techniques that affect the efficiency of the detection process. For instance, time and space consumption issue. As we know, both of the plagiarized ad reference documents are divided into n-grams before the comparison process. The process time of splitting the whole documents into n-grams and the space needed to store those n-grams will be a burden. In our proposed research, we will integrate the fingerprint matching technique with our proposed features based text similarity detection method. Rather than split the whole documents into n-gram, we will only choose the main ideas or important contents of the documents. Several features are considered to extract the main idea from the documents and integrated with the fingerprint matching technique for the comparison technique. All those proposed features are important elements to detect the plagiarized documents.

Below are four features that used by our proposed research to assist the plagiarism detection.

### 3.1 Top Keyword Feature

In order to detect translation plagiarisms, it is essential to translate the plagiarized Malay documents into English before used as the query documents for further detection process. After the plagiarized documents have been translated into English, it will improve the effectiveness of the detection process as the source documents are also in English. We use Google Translate API which is a well-known translation tool developed by the Google and is freely distributed. With this API, the language blocks of text can be easily detected and translated to other preferred languages. The API is designed to be a simple and easy to detect or translate languages when offline translation is not available.

### 3.2 First Sentence Similarity

In general, the first sentence of a document contains the important points of the overall document. The first sentence can act as a general extraction of the documents especially in news article, first sentence in the article is a very important sentence which can simply represent the overall contents of the article [4]. For this feature, we use the first sentence of the reference document to be compared with the plagiarized documents. Fingerprint matching technique is used to represent the sentence by dividing the first sentence and the plagiarized text into n-grams for comparison. The score for the first sentence similarity feature is computed by,

$$R = \frac{F(A) \cap F(B)}{F(A) \cup F(B)} \qquad 0 <= R <= 1$$

Where S(A) is the set of all n-grams in the first sentence of the reference document and S(B) is the set of all n-grams in the plagiarized document. . When the value of R is approaching 1, it means that the document is identical to the respective pair of document.

### 3.3 Query Phrase

In some condition, by taking the first sentence of an article to be the extraction or main ideas of the overall article, the effectiveness of the plagiarism detection will decreased. The first sentence similarity features is efficiently applied to most of the news articles, but not all types of the documents. Hence, we propose the query phrase feature that can be applied to all types of the documents. Generally, the query phase features implement the same flow of process with the first sentence similarity. The only difference is the sentence to be chosen as the extraction or main point of the document. Rather than taking the first sentence as the extraction, we choose the sentences which are considered important that appear after the query phrase. In this feature, we determine a few of the query phrases as below.

*In conclusion,*
*In general,*
*We conclude that…*
*We find that…*
*The survey shows that…*
*The experiment shows that…*

Based on this feature, the sentences that come after these query phrases are taken to be the extraction of the overall articles. These sentences can represent the overall ideas of the whole article effectively. For examples,

S1: We conclude that *the main cause of the social ills is the family problem.*
S2: In conclusion, *it cannot be denied that teachers play an important role in producing a group of good quality leader for the future.*

After determine the sentence, fingerprint matching technique is used to represent the sentence by dividing the chosen sentence and the plagiarized text into n-grams for comparison. The score for the first sentence similarity feature is computed by,

$$R = \frac{F(A) \cap F(B)}{F(A) \cup F(B)} \qquad 0 <= R <= 1$$

Where S(A) is the set of all n-grams in the chosen sentence of the reference document and S(B) is the set of all n-grams in the plagiarized document. . When the value of R is approaching 1, it means that the document is identical to the respective pair of document.

### 3.4 Longest Common Subsequence (LCS)

Longest Common Subsequence is one of the techniques used in ROUGE which is a well-known summary evaluation method. Given two sequences X and Y, the longest common subsequence (LCS) for X and Y is the common subsequence with maximum length [5]. To apply LCS in plagiarism detection, we view the documents as a sequence of words. The longer LCS of two documents sentences is, the more similar the two documents are. We use LCS-based F-measure to estimate the similarity between two documents X of length m and Y of length n, assuming X is a reference document sentences and Y is a plagiarize document.

$$R_{lcs} = \frac{LCS(X,Y)}{m}$$

$$P_{lcs} = \frac{LCS(X,Y)}{n}$$

$$F_{lcs} = \frac{(1+\beta\beta)R_{lcs}P_{lcs}}{R_{lcs}+(\beta\beta)P_{lcs}}$$

Where LCS(X,Y) is the length of a longest common subsequence of X and Y, and $\beta = P_{lcs}/R_{lcs}$. We notice that the score for the LCS is 1 when X=Y while score is zero when LCS(X,Y) =0 where there is nothing common in both between X and Y.

The main advantage of LCS is that it does not require consecutive matches but in-sequence matches that reflect sentence level word order as n-grams. The other advantage is that it automatically includes the longest in-sequence common n-grams, therefore no predefined n-gram length is necessary [5].

For example,

S1: *Player kicked the ball.*
S2: Player kick the ball.
S3: The ball kick player.

S1 is the reference sentence and S2 and S3 is two different plagiarized sentences respectively. If we implement the fingerprint matching technique with 2-gram, the score for S2 and S3 is the same since both have the same bigrams "the ball". However, S2 and S3 have significant difference in sentence meaning. When using the Longest Common Subsequence (LCS), S2 has a score of 3/4=0.75 and S3 has a score of 2/4=0.5, with $\beta = 1$. Hence, S2 has a higher similarity score than S3.

### 4 DISCUSSION

Feature based text similarity detection is a method to detect the plagiarism by integrating the use of fingerprint matching technique with several key features to assist the detection process. However, several aspects still need to be improved in order to increase the effectiveness and efficiency of the detection.

**Top Keyword Issue**

In general, after removal of stop words and stemming process, the word with higher frequency in a document can be considered as the key word that represent the overall ideas of the articles. However, some of the highly frequency words in the articles do not necessary provide the overall ideas of the document. This can be due to the author's style of writing who frequently writes the words that does not bring any significant meaning to the articles. In this situation, those words will be considered as top key words despite it does not provide any important meaning regards to the articles. The effectiveness of the plagiarism detection will affected indirectly due to this issue.

**First Sentence Similarity Issue**

As discussed in the previous section, the first sentence similarity issue is only effectively applied to particular types of articles especially the news articles. First sentence in most of the news articles contains the general ideas of the overall news. However, articles such as journal papers, thesis reports, and conference papers have the unique way to identify their most important sentence. First sentence of those types of articles do not provide the right idea on what the whole report is discuss about. Hence, by selecting the first sentence to be compared, the result of the detection cannot achieve its optimum level.

**Query Phrase**

Query Phrase feature is an effective solution to overcome the weaknesses in the first sentence similarity feature. It provides a different way to capture the important sentence in an article and take those sentences to be compared using the fingerprint matching technique. In our proposed research, we determine 6 major query phrases to be used in the method. However, there are still a lot of phrase can be considered as query phrase which can provide a more effective plagiarism detection process. Future works should be done to increase the number of suitable query phrase that used in out proposed method.

### 5 CONCLUSION

In this paper, we presented our method in text similarity detection. Feature based text similarity detection is a new approach to detect plagiarism. This approach integrates the use of fingerprint matching technique with several key features to detect the similarity level between sentences. With further experiment, we feel that our approach can detect the similarity between sentences with high efficiency and effectiveness.

## 6  ACKNOWLEDGEMENT

This project is sponsored partly by the Ministry of Science, Technology and Innovation under the E-Science grant 01-01-06-SF-0502.

.

**Mr. Chow Kok Kent** received his B.Sc. degree in Universiti Teknologi Malaysia, Malaysia in 2009. He is currently purchasing Master degree in Faculty of Computer Science and Information System, Universiti Teknologi Malaysia. His current research interest includes information retrieval, Plagiarism Detection and Soft Computting.

**Dr. Naomie Salim** is an Assoc.Prof. presently working as a Deputy Dean of Postgraduate Studies in the Faculty of Computer Science and Information System in Universiti Teknologi Malaysia. She received her degree in Computer Science from Universiti Teknologi Malaysia in 1989. She received her Master degree from University of Illinois and Ph.D Degree from University of Sheffield in 1992 and 2002 respectively. Her current research interest includes Information Retrieval, Distributed Database and Chemoinformatic.

.